\documentclass[conference]{IEEEtran}
\IEEEoverridecommandlockouts
\usepackage{cite}
\usepackage{amsmath,amssymb,amsfonts}
\usepackage{algorithmic}
\usepackage{graphicx}
\usepackage{url}
\usepackage{pdfpages}
\usepackage{textcomp}
\usepackage{xcolor}
\usepackage{graphicx} % Para inserir imagens como PNG, JPG ou PDF
\usepackage{hyperref} % Para links clicáveis
\def\BibTeX{{\rm B\kern-.05em{\sc i\kern-.025em b}\kern-.08em
    T\kern-.1667em\lower.7ex\hbox{E}\kern-.125emX}}

\newcommand{\orcid}[1]{~\href{https://orcid.org/#1}{\includegraphics[trim=0 0 0 0,clip,width=9pt,height=9pt]{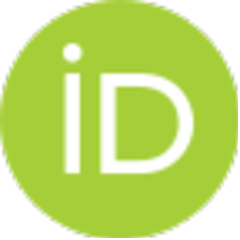}}}

\begin{document}

\title{
%Bayesian Graph Neural Networks for\\Event Sifting with Human-in-the-loop
Interactive Event Sifting using Bayesian Graph Neural Networks
}

\author{
\IEEEauthorblockN{José Nascimento\orcid{0000-0003-3450-6029}}
\IEEEauthorblockA{Artificial Intelligence Lab., \url{Recod.ai} \\
\textit{Univ. of Campinas (Unicamp)}\\
Campinas, SP, Brazil \\
\url{jose.nascimento@ic.unicamp.br}}
\and
\IEEEauthorblockN{Nathan Jacobs\orcid{0000-0002-4242-8967}}
\IEEEauthorblockA{Multimodal Vision Research Lab \\
\textit{Washington University in St. Louis}\\
Saint Louis, MO, United States \\
\url{jacobsn@wustl.edu}}
\and
\IEEEauthorblockN{Anderson Rocha\orcid{0000-0002-4236-8212}}
\IEEEauthorblockA{Artificial Intelligence Lab., \url{Recod.ai} \\
\textit{Univ. of Campinas (Unicamp)}\\
Campinas, SP, Brazil \\
\url{arrocha@unicamp.br}}

}

\maketitle

\begin{abstract}

Forensic analysts often use social media imagery and texts to understand important events. A primary challenge is the initial sifting of irrelevant posts. This work introduces an interactive process for training an event-centric, learning-based multimodal classification model that automates sanitization. We propose a method based on Bayesian Graph Neural Networks (BGNNs) and evaluate active learning and pseudo-labeling formulations to reduce the number of posts the analyst must manually annotate. Our results indicate that BGNNs are useful for social-media data sifting for forensics investigations of events of interest, the value of active learning and pseudo-labeling varies based on the setting, and incorporating unlabelled data from other events improves performance.

\end{abstract}

\begin{IEEEkeywords}
Bayesian Graph Neural Networks, forensic event analysis, human-in-the-loop, few-shot learning
\end{IEEEkeywords}

\section{Introduction}

Social media posts contain imagery and texts that give relevant insights to analysts in a forensic event of interest. Nevertheless, automatically sifting the informative data from the myriad of irrelevant content retrieved after a keyword-based search is necessary to reduce the workload of forensic experts. The use of machine learning in this scenario is a promising way forward. However, one bottleneck is that established deep-learning-based multimedia classification typically requires strong annotation to perform well, which requires time and human work that might not be available in a real scenario.

We envision a classification system for forensic events operated by an analyst in real time. This analyst should be integral to the decision loop, selecting the first annotated instances and carrying out active learning. This approach ensures that the method classifies each post and identifies those most crucial for improving the model in the next iteration, improving final decision-making in the forensic investigation.

Developing a classification model for this purpose presents two significant challenges. Firstly, the model must deliver high performance with limited annotated data points, as we need to minimize human-required efforts in the active learning process. Secondly, it should accurately measure classification uncertainty, a key requirement for active learning. We rely upon a Bayesian Graph Neural Network to address these challenges. This choice is informed by the strong performance of GNNs in the few-shot learning setting and the ability of Bayesian Neural Networks to provide a comprehensive understanding of a given instance's prediction for a posterior active learning method. Additionally, we have enhanced our formulation with clustering for instance selection, dataset augmentation, and pseudo-labeling.
Therefore, the key contributions of this work are threefold:
\begin{enumerate}
    \item We adapted MC Dropout (Monte Carlo Dropout) to be a Bayesian component of a Graph Neural Network and demonstrate its improvement in the Active Learning setting.
    \item For the human-in-the-loop part, we extend upon BALD (Bayesian Active Learning by Disagreement) to consider diversity when selecting annotated instances. We also rely upon KMeans for data selection in the first iteration.
    \item We leverage unlabeled data from other events allied with BALD for pseudo-labeling to enrich diversity and quality of response.
\end{enumerate}

\section{Related Work}\label{sec:lit-review}

Previous works on multimodal event sifting focused on how to combine different modalities to separate informative and non-informative data. Techniques used in this sense include global attention~\cite{susladkar2023}, cross attention~\cite{Liang2022}\cite{qian2023}, co-attention~\cite{LIN2024103697}, constrastive learning~\cite{mandal2024}, and score-level fusion~\cite{YOON2023101922}. However, works for this task focusing on scarce labels are rare. Some methods train improved semi-supervised methods such as FixMatch~\cite{sirbu-etal-2022-multimodal}, and MixMatch~\cite{abavisani2020} to work around the absence of annotation, whereas other methods use additional information such as annotated data from similar events~\cite{abavisani2020}. However, in a real-world scenario, training these semi-supervised models from scratch takes time and resources, and it is not guaranteed to have a labeled dataset for a similar event.

% In the forensic field, in general, the use of human-in-the-loop techniques is uncommon. Some tasks it has been experimented with include fact-checking~\cite{yang2022} and vulnerability detection~\cite{nguyen2021}. 
% For event sifting, we found works with instance selection~\cite{nascimento2022} and active learning~\cite{matthieu2023}. Nevertheless, the first work does not propose an iterative pipeline, and the second is based on uncertainty measures, which might not be reliable in the deep learning context. 

Human-in-the-loop approaches have been proposed for other forensic domains, including fact checking~\cite{yang2022} and vulnerability detection~\cite{nguyen2021}.
For multimodal event sifting, instance selection~\cite{nascimento2022} has been explored, but not along with an iterative pipeline. In turn, active learning~\cite{matthieu2023} also has been explored, but solely based on uncertainty measures of the softmax output of a neural network, which might not be reliable for this purpose.

\section{Proposed Method}

\begin{figure*}
  \centering
  \includegraphics[width=\textwidth]{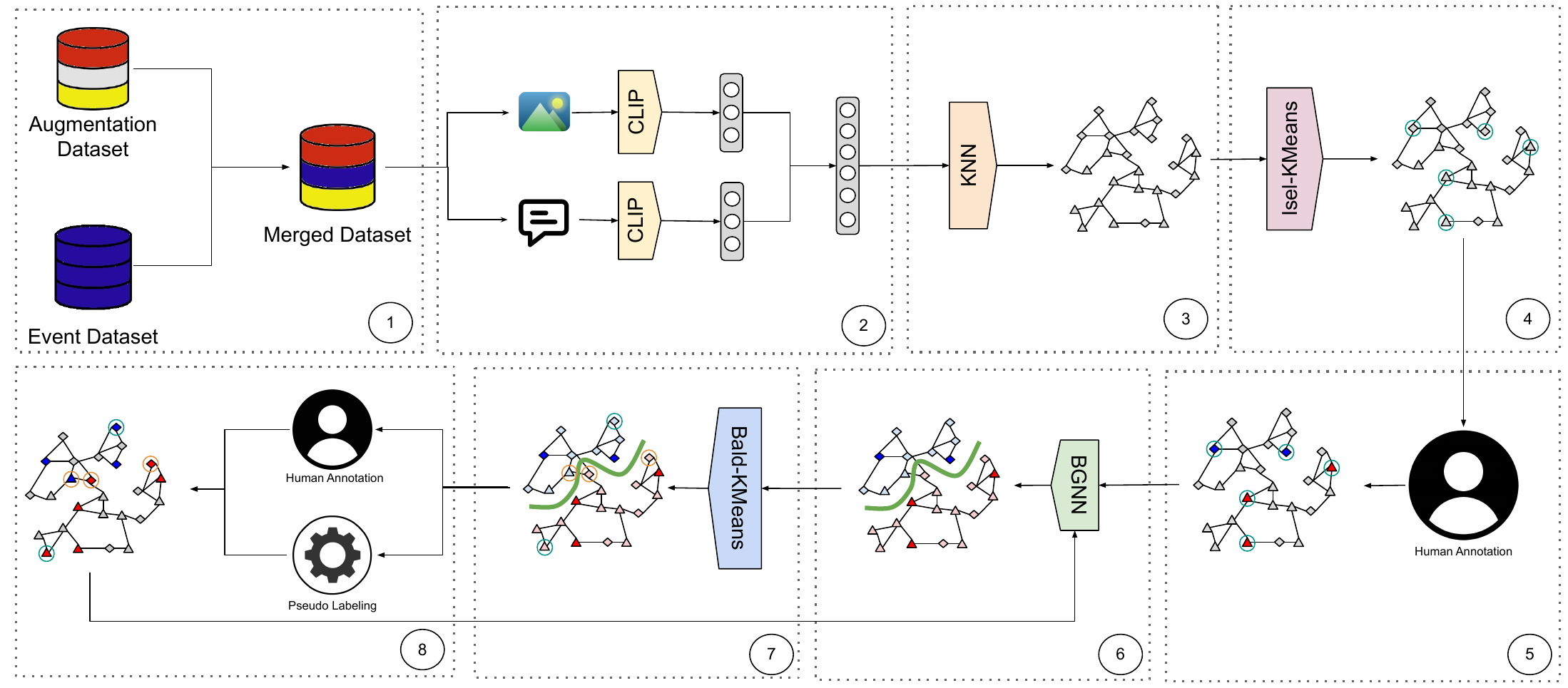}
  \caption{\textbf{Proposed Method}. In (1), we combine an event dataset of interest with the augmentation dataset, which combines data from other events. All augmentation dataset instances are considered negative samples during training. In (2), we extract features from the image and text attached to the post using a CLIP model and concatenate them to form a post representation, and in (3), we use k-NN to support the graph creation. In (4), we run KMeans and select the point closer to each centroid to be annotated by the user (the analyst in a hypothetical event analysis scenario). With some points annotated in (5), we run the Bayesian Graph Neural Network in (6), obtaining suitable predictions for the next step (7), where BALD-KMeans measures the most certain and least certain points in each cluster. Finally, in (8), the analyst provides the labels for the points with the most uncertainty. In addition, the method applies pseudo-labeling in certain cases, considering the model prediction as ground truth. Finally, the method goes back to step (6) for another iteration.} 
  \label{fig:pipeline}
\end{figure*}

%In the method presented in this paper, we iteratively classify multimodal data regarding whether or not they are related to a forensic event of interest. 
We propose a human-in-the-loop algorithm for classifying whether a social media post is related to a forensic event of interest.
%A broad overview of the pipeline is provided in Figure~\ref{fig:pipeline}, and the details of each step are described next.
Figure~\ref{fig:pipeline} provides an overview of the architecture, and further details are explained in the remainder of this section. In addition, the source code\footnote{\url{https://github.com/jdnascim/bnn-al}} for the experiments is publicly available.

\subsection{Event Augmentation}

For diversity and data representativeness, 
%representativity
we add posts from other events to the training set, automatically labeling it as a third class (regardless of its content, which makes it different from prior work~\cite{abavisani2020}, for instance, as it does not rely on annotations). 
Data from other events can be easily obtained from existing datasets available in the literature.
This strategy enhances the overall performance of the classification method and the human-in-the-loop part, and we hypothesize that the irrelevant data from other events might share similarities with irrelevant data from the event in such a way that the relevant data does not. We ensure that sampled data is only from events whose ``type'' differs from the event of interest. 
%In other words, if we want to classify the ``Mexico Earthquake'', we consider every event but the ``Iran-Iraq Earthquake'' in the augmentation set. 
Besides, we limited the size of this additional set to the size of the original event dataset.

During training, the labeled set includes data annotated as ``informative,'' ``not informative,'' and ``another event.'' In contrast, the unlabeled and test sets only include ``informative'' and ``not informative'' labels. In inference, we treat ``another event'' as ``not informative''. This approach aligns with our use of a balanced loss function, as it ensures a higher balance of "not informative'' data from the event-of-interest dataset in the loss calculation.

\subsection{Feature Extraction, Fusion, and Graph Construction}

Given that we only consider posts containing images and text, we opted for a CLIP model based on a ViT-B/32 Transformer architecture for feature extraction~\cite{clip2021}. CLIP models target representing images and texts in a shared vector space and are widely used in multimodal scenarios, given that they have presented good results in different tasks. After the feature extraction, we concatenate both vectors and consider these features to represent the posts.

%\subsection{Graph Construction}
After that,
we need to build a graph connecting similar points to use graph-based algorithms. We rely upon the k-Nearest Neighbors (k-NN) algorithm, using the cosine distance as a similarity measure. We adopted $k = 16$ to ensure the graph remains sparse, optimizing efficiency.

\subsection{Instance Selection}

We need some annotated data for the first iteration. Nevertheless, we still lack model uncertainties. To address this, we use the KMeans clustering algorithm. For the sake of simplicity, we defined $18$ centroids, matching the number of instances we plan to annotate at this step. After running the clustering algorithm, we select the closest point to each of the $18$ centroids for labeling.

\subsection{Bayesian Graph Neural Network and BALD}~\label{sec:bayesian}

Graph Neural Networks have been well studied in the semi-supervised context and used in different works for this task. In summary, it works around the absence of labels by incorporating unlabeled data information into labeled data through the edges of the graph. In addition, Bayesian Networks are important in Active Learning, given that previous studies indicate that regular deep networks' logits might not be reliable for uncertainty measures~\cite{guo2017}, something paramount for Active Learning.

The GNN consists of two GraphSAGE (Graph Sample and Aggregate)~\cite{NIPS2017_5dd9db5e} layers (one of size 1024, and the other of size 2048), and one linear layer for classification. GraphSAGE, a well-established GNN layer, integrates neighborhood information with learnable parameters, making it ideal for inductive learning. To have a Bayesian GNN (henceforth called BGNN), we used MC Dropout~\cite{pmlr-v48-gal16}. This technique simulates the Bayesian prediction by running inference $n$ times with dropout turned on. With numerous different scores for the same model, we can go to the next step. In this paper, we opted for $n = 10$.

% \subsection{Uncertainty Measurement}

At this stage, we have multiple predictions for the same point. To measure the uncertainty, we calculate the BALD score~\cite{Houlsby2011BayesianAL}. Given a prediction $p_{CK}$ for $C$ classes and $K$ different scores for a given instance, the BALD score is calculated by:
\begin{equation}
    BALD(p_{CK}) =  \frac{1}{K} \sum^{C}_i \sum^{K}_j p_{ij} * e^{p_{ij}} - \sum^{C}_i \bar{p_{i}} * e^{\bar{{p_{i}}}} \label{eq:bald}
\end{equation}
where:
\begin{equation}
    \bar{p_{i}} = ln(\frac{1}{K} \sum^{K}_{j} e^{p_{ij}}).
\end{equation}
The first part of Equation~\ref{eq:bald} measures the conditional entropy and scores high values for higher predictions (regardless of consistency), whereas the second part measures the entropy and tends to give a prediction closer to the first part but penalizes high and inconsistent predictions due to the mean factor ($\bar{pi}$). Given that the entropy signal is negative, items with high and inconsistent predictions will have the highest BALD score.

We chose other hyperparameters through a tuning procedure, specifically dropout with $p = 0.5$, $500$ epochs, $1\mathrm{e}{-5}$ for the learning rate, $1\mathrm{e}{-2}$ for the weight decay. We used \textit{Negative Log Likelihood} as a loss.

\subsection{Active Learning (analyst interactions)}

At this stage, we have initial predictions and their uncertainties. The next step consists of selecting instances to be annotated for the next iteration. Items with higher BALD scores are preferable here. Nevertheless, only using BALD raises a possible problem: the need for more diversity. We infer that an instance might have a BALD score closer to its neighborhood. That might be a problem because only selecting the instance with the highest score might lead to similar data being selected. In this perspective, we first run KMeans with $n_{c} = 16$; then, we discard all clusters with less than $16$ items (to prioritize items in a higher dense space) and select the instance with the highest BALD score to be annotated in each cluster. If one or more clusters were discarded, we select enough highest score items from each cluster to be annotated and complete $16$ annotations. If $16$ is not divisible by the number of remaining clusters, we randomly select the groups needing one more labeled instance than the others.

\subsection{Pseudo-Labeling}

Given an instance with high and consistent prediction, it will have a small BALD score, given that its entropy and conditional entropy will be close. Because of that, we infer that automatically labeling the most certain instances with their predictions might also improve the final classification performance of an event of interest. To do so, we rely on the same clustering from the active learning step to select the $16$ instances with the lowest BALD of each group, as this lower score indicates higher confidence.

\section{Experiments and Results}

In this section, we describe the dataset and then present an ablation test and a comparison with other models, highlighting how each part of the method improves the performance.
We used a simulated environment with dataset ground truth rather than conducting experiments with humans, as it is a common procedure in the active learning literature~\cite{10.3389/frai.2022.737363}.

\subsection{Dataset}

In this work, we adopt CrisisMMD dataset~\cite{crisismmd2018icwsm}, an established dataset in the crisis informatics scenario. It consists of seven events of different types (hurricanes, wildfires, earthquakes, and floods) and diverse sizes (from 500 posts to 4k).  Table~\ref{tab:dataset} shows the events of interest in this dataset. We also considered only posts with images attached to them and ruled out duplicates.   
In this paper, we run each experiment with ten different random seeds for each of the seven events separately (resulting in 70 experiments), and then we averaged the results in the test set across the datasets. Our method, based on semi-supervised learning, ensures the training set is comprehensive, consisting of both labeled and unlabeled data. This thorough approach, even though the training set remains the same, allows for potential changes in the labeled instances according to the seed, as long as there is some random process in the instance selection, which is the case for KMeans. 
% By running ten different random seeds in seven different datasets, each experiment consists of 70 different runs, demonstrating the depth of our research.

        \begin{table}[ht]
    \centering
    \caption{Seven events which consist of the CrisisMMD datasets.}
    \begin{tabular}{|ll|rr|}
    \hline
    \textbf{Event Name}  & \textbf{Type} & \textbf{Train Set Size} & \textbf{Test Set Size} \\ \hline
    Hurricane Irma       & Hurricane    & 3,421                    & 535                    \\
    Hurricane Harvey     & Hurricane    & 3,285                    & 568                    \\
    Hurricane Maria      & Hurricane    & 3,462                    & 552                    \\
    California Wildfires & Wildfires    & 1,139                    & 210                    \\
    Mexico Earthquake    & Earthquake   & 1,044                    & 177                    \\
    Iraq-Iran Earthquake & Earthquake   & 452                     & 72                     \\
    Sri Lanka Floods     & Floods       & 805                     & 110                    \\ \hline
    \end{tabular}
    \label{tab:dataset}
    \end{table}

\begin{table*}[!htbp]
\centering
\caption{Ablation test for the main components of our proposed method.}

\begin{tabular}{|cccc|llll|}
\hline
\multicolumn{4}{|c|}{\textbf{Parameter}}                                                                                                               & \multicolumn{4}{c|}{\textbf{F1-Score (\%)}}                                                                                    \\ \hline
\multicolumn{1}{|c|}{\textbf{Augmentation Dataset}} & \multicolumn{1}{c|}{\textbf{Instance Selection}} & \multicolumn{1}{c|}{\textbf{Active Learning}} &\textbf{Pseudo-Labeling} & \multicolumn{1}{c|}{\textbf{18}} & \multicolumn{1}{c|}{\textbf{34}} & \multicolumn{1}{c|}{\textbf{50}}                & \multicolumn{1}{c|}{\textbf{Sum}} \\ \hline
no                                         & random                                  & random                               & no              & 69.0±7.0                & 73.3±6.1                & \multicolumn{1}{l|}{76.0±4.2}          & 218.3±17.3               \\ \hline
no                                        & random                                  & random                          & \textbf{yes}              & 69.0±7.0       & 76.5±3.0                & \multicolumn{1}{l|}{77.8±1.7}          & 223.3±11.7               \\
no                                        & random                                  & \textbf{bald-kmeans}                               & no             & 69.0±7.0       & 74.2±5.0                & \multicolumn{1}{l|}{77.1±2.6}          & 220.3±14.6               \\
no                                        & \textbf{kmeans}                                  & random                          & no             & 72.4±5.7               &  76.3±2.9                & \multicolumn{1}{l|}{77.1±2.5} & 225.5±11.1               \\
\textbf{yes}                                         & random                                  & random                          & no   & 70.6±8.5                & 75.6±2.6                & \multicolumn{1}{l|}{76.6±2.8}          & 222.8±13.9               \\ \hline
yes                                        & kmeans                                  & bald-kmeans                          & \textbf{no}              & \textbf{73.6±5.2}       & 75.5±3.4                & \multicolumn{1}{l|}{76.8±2.6}          & 225.9±11.2               \\
yes                                        & kmeans                                  & \textbf{random}                               & yes             & \textbf{73.6±5.2}       & 76.6±2.5                & \multicolumn{1}{l|}{77.3±2.2}          & 227.5±9.9               \\
yes                                        & \textbf{random}                                  & bald-kmeans                          & yes             & 70.6±8.5               & 76.8±3.1                & \multicolumn{1}{l|}{\textbf{78.1±1.9}} & 225.5±13.5               \\
\textbf{no}                                         & kmeans                                  & bald-kmeans                          & yes             & 72.4±5.7                & 76.4±3.4                & \multicolumn{1}{l|}{77.9±2.3}          & 226.7±11.4               \\ \hline
yes                                        & kmeans                                  & bald-kmeans                          & yes             & \textbf{73.6±5.2}       & \textbf{77.0±2.5}       & \multicolumn{1}{l|}{77.8±2.1}          & \textbf{228.4±9.8}      \\ \hline
\end{tabular}
\label{tab:ablation}
\end{table*}

\subsection{Ablation Test}

\begin{figure}
  \raggedright
  \includegraphics[width=0.5\textwidth]{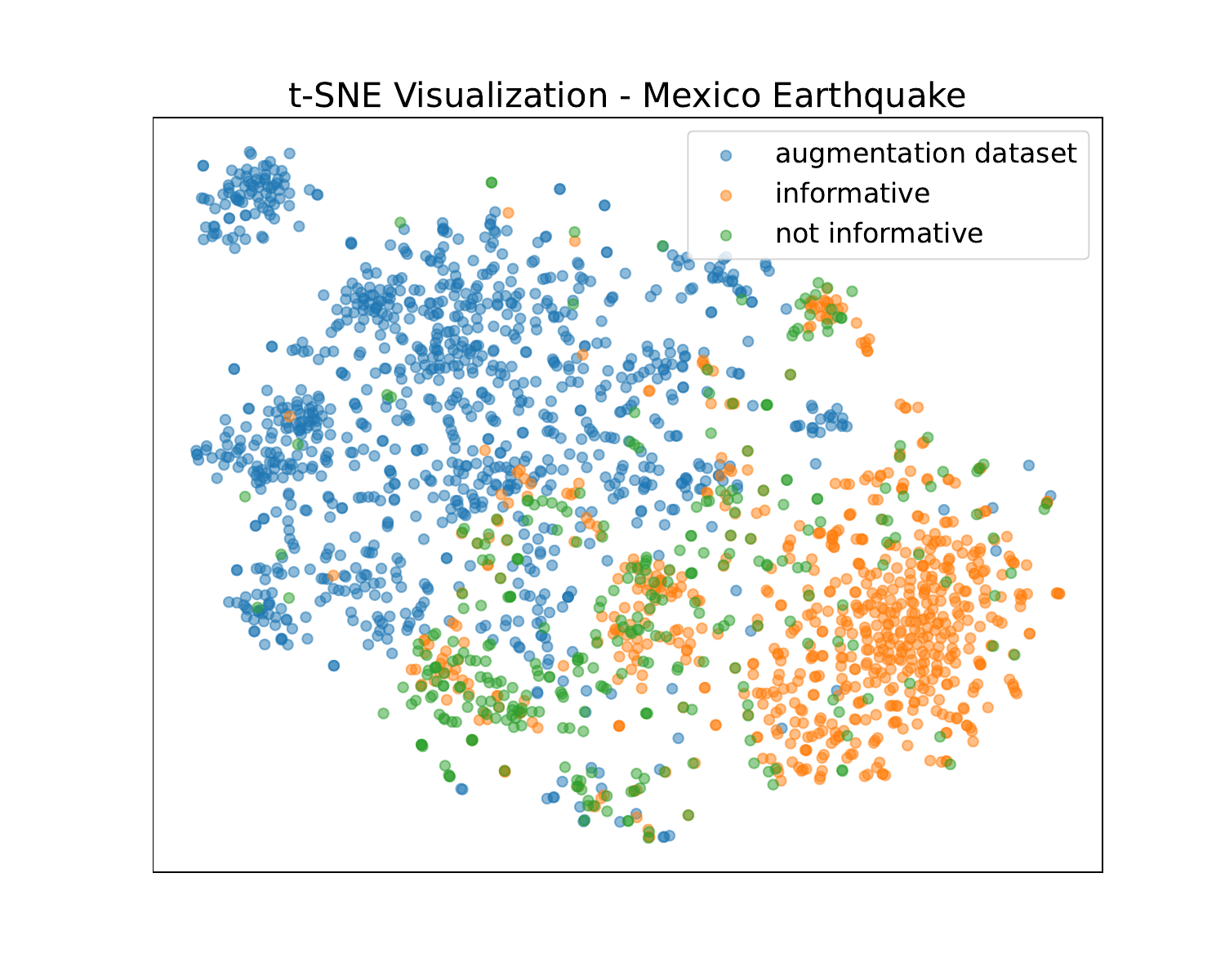}
  \caption{\textbf{t-SNE Visualization}. 2D visualization for the Mexico Earthquake dataset. The data distribution of the augmentation dataset is closer to the sparse non-informative set, which we infer to be the reason for the improvement caused by its addition to the graph.} 
  \label{fig:tsne-vis}
\end{figure}

In Table~\ref{tab:ablation}, we present the impact of each piece of our method on the overall task of event sifting,
in a context where we start with 18 annotated instances chosen by instance selection, then run active learning to obtain 16 more instances twice, creating steps with 34 and 50 labels. For 18 labeled items, we do not have prior predictions; therefore, only the use of the augmentation dataset and the instance selection method impact the results. We can see that using both positively influenced this first iteration, with an improvement of almost four percentage points compared to full random selection without augmentation. The better performance using the augmentation dataset might indicate some pattern regarding the data distribution. We hypothesize that non-related information is more sparse in the vectorial space and that additional data from other events create denser clusters around at least part of it. We present a visualization of the data in Figure~\ref{fig:tsne-vis}. We can see that informative data are more clustered than not informative and that the augmentation dataset tends to be closer to the non-informative subset.

For 34 labeled instances, the impact of the instance selection method seems to decrease, and the use of active learning and pseudo-labeling is more important. The considerable improvement caused by pseudo-labeling indicates that only relying on the most uncertain data might not be the best choice. 
%We hypothesize that the most certain ones are also more influential, and adding them to the loss might improve the overall result because their information will be more spread on the graph.

Finally, for 50 labeled instances, the tendency from the previous iteration remains, with a disclaimer that the gaps between the different configurations are shorter. This indicates that the methods experimented in this paper are more decisive in the few-shot learning scenario. In other words, they work when just a few annotations are available, which is exactly the key objective of this research.

\subsection{Model Comparison}

Table~\ref{tab:baynet} compares our Bayesian Graph Neural Network (BGNN) with different models of similar size: a Bayesian Multilayer Neural Network (BNN), a regular Graph Neural Network (GNN), and a Multilayer Neural Network (MLP). For the last two, we used uncertainty with KMeans instead of the BALD score for active learning, given that BALD is unsuitable for non-bayesian networks. The other parts of the pipeline remained the same. 
%The proposed model presents the best results, and the good results for BMLP indicate the advantage of using the Bayesian setup.
The results suggest the superiority of GNNs and Bayesian networks for this task, as we mentioned in Section~\ref{sec:bayesian}, and that this superiority holds when we combine both techniques.

\begin{table}[]
\centering
\caption{Performance comparison of different models with BGNN}
\begin{tabular}{|cc|ccc|}
\hline
\multicolumn{2}{|c|}{\textbf{Parameter}}                                                                                                               & \multicolumn{3}{c|}{\textbf{F1-Score (\%)}}                                                                                    \\ \hline
\multicolumn{1}{|c|}{\textbf{Model}} & \textbf{AL} & \multicolumn{1}{c|}{\textbf{18}} & \multicolumn{1}{c|}{\textbf{34}} & \textbf{50}                   \\ \hline

MLP            & unc-kmeans  & 70.0±7.4                         & 75.5±3.5                         & 76.9±2.7                      \\ 
GNN            & unc-kmeans  & \textbf{74.0±3.2}                         & 76.4±2.9                         & 77.4±2.8                      \\
BMLP           & bald-kmeans & 70.0±6.9                         & \multicolumn{1}{l}{76.6±3.1}     & \multicolumn{1}{l|}{\textbf{77.8±2.9}} \\ \hline
BGNN           & bald-kmeans & 73.6±5.2                         & \multicolumn{1}{l}{\textbf{77.0±2.5}}     & \multicolumn{1}{l|}{\textbf{77.8±2.1}} \\ 
\hline
\end{tabular}
\label{tab:baynet}
\end{table}

\subsection{Runtime}

Given that the method is intended for interactive use with humans, achieving good efficiency is paramount. We trained our model on an Nvidia RTX 6000, with an average runtime of 117 ± 84 seconds for the first iteration, which involved model training, feature extraction, and graph spanning, based on ten runs with different seeds across seven datasets. The second and third iterations, which included both model training and BALD calculation, had an average runtime of 121 ± 72 seconds, using the same datasets and seeds as in the first iteration.

\section{Conclusion and Future Work}

Relying upon human-in-the-loop techniques in the forensics setup might be important for a task needing more training labels, like event sifting. This work suggests using a Bayesian Graph Neural Network and BALD-based active learning instead of regular neural networks using the consolidated uncertainty measure. Also, it suggests pseudo-labeling and the addition of data from previous events to improve performance, as well as KMeans from the first data selection, when we do not have BALD predictions. In summary, we present a full iterative pipeline from the first selected data points to be annotated to the final classification and decision-making. 

For future work, more graph features for this setting might be researched. To illustrate, a community algorithm might be experimented with to replace KMeans, and node features for influence, such as betweenness and degree, might be explored to choose data to be annotated. Also, an improved way to construct the graph might impact the final result, such as adding edge features or using a heterogenous graph. On the other hand, improvements in runtime are desirable. One possible path of research is an alternative method to construct the graph for larger datasets, as constructing a similarity matrix is costly. Finally, measuring the performance of the method presented in this work with humans might identify potential adjustments necessary in a real-world scenario.

% Additional graph features could be researched for future work, such as using a community algorithm instead of KMeans and incorporating influence metrics like betweenness and degree for selecting data to annotate. Improvements in graph construction, such as adding edge features or using heterogeneous graphs, might also impact the results. Alternative methods for constructing graphs in larger datasets could be investigated due to the cost of creating similarity matrices. Lastly, testing the technique with human input could reveal necessary adjustments in a real-world scenario.

\section{Acknowledgements}

We thank the McDonnell International Scholars Academy at Washington University in St. Louis and the São Paulo Research Foundation (FAPESP) Horus project (Grant \#2023/12865-8) for supporting this work.

\bibliographystyle{IEEEtran}
\bibliography{bibliography}

\end{document}